\documentclass[11pt,a4paper]{article}
\usepackage{authblk}
\usepackage[hyperref]{eacl2021}
\usepackage{times}
\usepackage{latexsym,booktabs,comment}
\usepackage{amssymb}
\usepackage{graphicx}
\usepackage{url}
\usepackage{tabularx}
\usepackage{todonotes}
\usepackage{xr}

\aclfinalcopy 
\def\aclpaperid{331} 

\makeatletter
\newcommand*{\addFileDependency}[1]{
  \typeout{(#1)}
  \@addtofilelist{#1}
  \IfFileExists{#1}{}{\typeout{No file #1.}}
}
\makeatother

\newcommand*{\myexternaldocument}[1]{%
    \externaldocument{#1}%
    \addFileDependency{#1.tex}%
    \addFileDependency{#1.aux}%
}
\myexternaldocument{appendix}

\newcommand{\bcub}{B$^3$}
\newcommand{\ceaf}{CEAF$_{\phi_4}$}

\title{Ellipsis Resolution as Question Answering: An Evaluation}

\author[1]{\textbf{Rahul Aralikatte}}
\author[1,2]{\textbf{Matthew Lamm}}
\author[3]{\textbf{Daniel Hardt}}
\author[1]{\textbf{Anders S{\o}gaard}}
\affil[ ]{$^1$University of Copenhagen, $^2$Stanford University} \affil[ ]{$^3$Copenhagen Business School}

\graphicspath{{./images/}}

\begin{document}
\maketitle
\begin{abstract}
Most, if not all forms of ellipsis (e.g., `{\em so does} Mary') are similar to reading comprehension questions (`{\em what does} Mary do'), in that in order to resolve them, we need to identify an appropriate text span in the preceding discourse. Following this observation, we present an alternative approach for English ellipsis resolution relying on architectures developed for question answering (QA). We present both single-task models, and joint models trained on auxiliary QA and coreference resolution datasets, clearly outperforming the current state of the art for Sluice Ellipsis (from 70.00 to 86.01 F$_1$) and Verb Phrase Ellipsis (from 72.89 to 78.66 F$_1$).  
\end{abstract}

\section{Introduction}
Ellipsis resolution is a hard, open problem in NLP, and an important source of error in machine translation, question answering, and dialogue understanding \cite{Vicedo:Ferrandez:00,Dzikovska:ea:06,Chung:Gildea:10,Macketanz:ea:18,Bach:ea:20}. 
There are no large annotated text corpora for this phenomenon, even for English, and we only have annotations for a subset of the known ellipsis constructions. Since annotation is expensive and cumbersome, any synergies with existing NLP tasks could be useful and enable us to leverage auxiliary data when learning models for ellipsis resolution.








\begin{figure}
    \centering
    \includegraphics[width=\columnwidth]{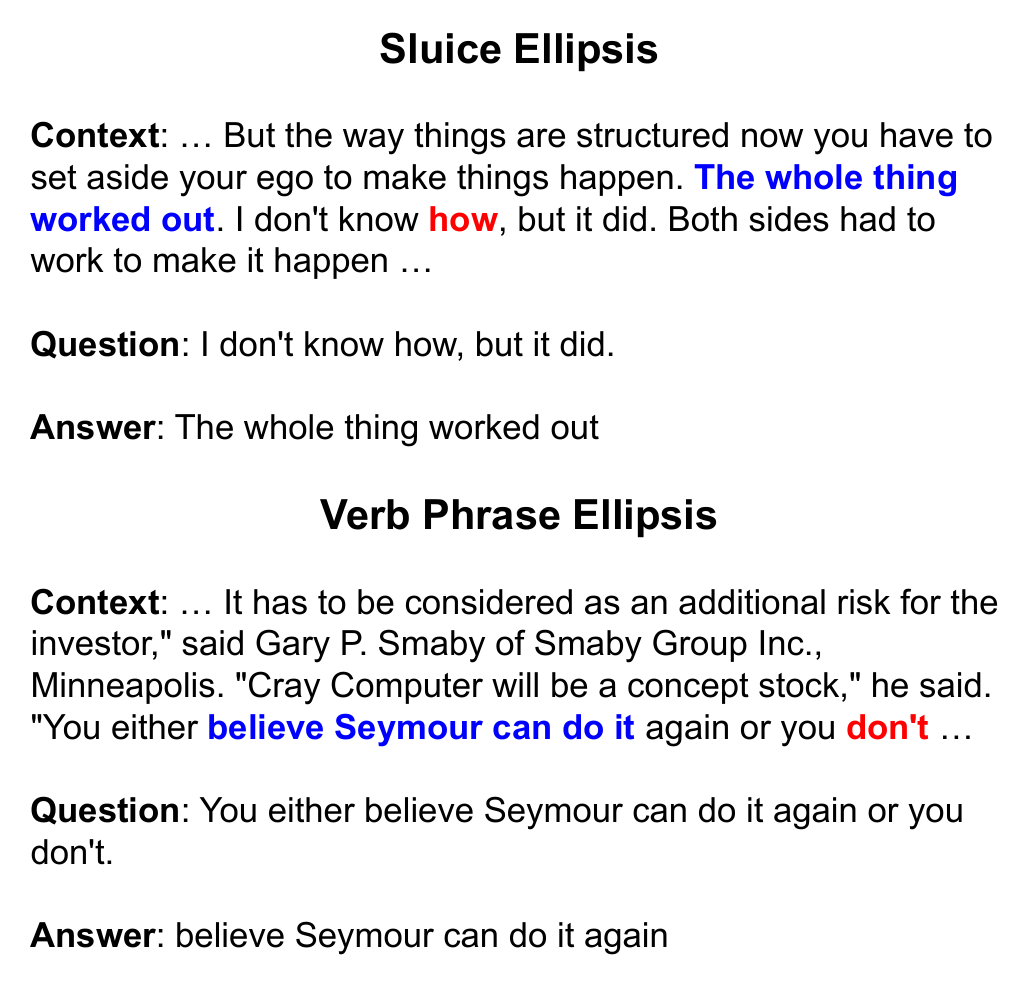}
    \caption{Examples of Sluice Ellipsis and Verb Phrase Ellipsis, represented as ``questions" about their associated contexts. Wh-phrases and auxiliary verbs are marked in \textcolor{red}{red} and elided phrases are marked in \textcolor{blue}{blue}.} 
    \label{fig:qa_example}
\end{figure}

This paper presents a simple yet strong approach to ellipsis resolution based on a straightforward observation, depicted in Figure \ref{fig:qa_example}, that ellipsis resolution can be converted to a QA problem. 
Ellipsis and questions put in focus {\em referentially dependent} expressions \cite{Carlson:06}, or free variables \cite{Partee:78}, that need to be resolved in order to comprehend the discourse. For similar observations about different tasks, see \newcite{McCann:ea:18} and \newcite{Gardner:ea:19}.


\noindent This straightforward observation leads us to suggest treating different forms of ellipsis resolution -- and later, as an auxiliary task, coreference resolution -- as a QA problem, and to apply state-of-the-art architectures for QA to ellipsis resolution tasks, as well as to experiment with using training data for QA and coreference resolution to improve our new ellipsis resolution models.

\paragraph{Contributions} We cast ellipsis as a QA problem, enabling us to induce models for it using neural architectures originally developed for QA. Applying these architectures out of the box enables us to establish strong results\footnote{Though we report state-of-the-art results for both sluice and verb phrase ellipsis, we consider these models as strong baselines for future research as they are obtained purely using existing methods.} for ellipsis resolution tasks, improving significantly over previous work. Using the same architecture for the different ellipsis resolution tasks, as well as for QA and coreference resolution, enables us to explore synergies between the tasks, and we show that training joint models on these tasks leads to even better performance.

\section{Methodology}

In this section, we briefly describe the various datasets used for training, and explain how they are converted into QA format. We then move on to the choice of model architectures and the reasoning behind their selection.

\paragraph{Sluice Ellipsis}
For training and evaluation of Sluice Ellipsis resolution models, we use the corpus introduced by \newcite{anand-esc}, which contains 3,103 annotated examples of embedded sluices, collected from the New York Times section of the English Gigaword corpus. Since the annotators were free to paraphrase the antecedent, in some cases, a string match on the context does not return antecedent span indices. To ensure a fair comparison, we follow previous work \cite{sluice-short}, which is also the current state-of-the-art, in ignoring these instances, and use their split for training, development and testing.

\paragraph{Verb Phrase Ellipsis}
\newcite{bos-vpe} provide Verb Phrase (VP) Ellipsis annotations for the WSJ part of the Penn Treebank. All $25$ sections were annotated, and we follow them in using sections 0-19 for training, and 20-24 for testing. We further hold out sections 18-19 from the training data for development. This also enables to us compare our results directly with the current state-of-the-art for VP Ellipsis \cite{zhang-2019}.

\paragraph{Coreference Resolution}
For coreference resolution, which we use as an auxiliary task, we train and evaluate on two corpora: (i) the English portion of the OntoNotes 5.0\footnote{\url{https://catalog.ldc.upenn.edu/LDC2013T19}} corpus with the standard data split used in the CoNLL-2012 shared task \cite{conll2012}, and (ii) the WikiCoref corpus \cite{wikicoref}, which contains annotations of $30$ documents from the English Wikipedia. From this dataset, we use $22$ documents for training, $4$ documents for development, and $4$ for testing.

\paragraph{QA} We also use SQuAD v1.1 \cite{squad} as an auxiliary reading comprehension dataset.

\begin{table}
    \centering
    \small
    \begin{tabular}{>{\raggedleft\arraybackslash}m{2.3cm}|
                    >{\raggedleft\arraybackslash}m{0.8cm}
                    >{\raggedleft\arraybackslash}m{0.8cm}
                    >{\raggedleft\arraybackslash}m{0.8cm}
                    >{\raggedleft\arraybackslash}m{0.8cm}}
    \toprule
    \textbf{Task} & \textbf{Train} & \textbf{Dev} & \textbf{Test} & \textbf{ACL} \\
    \midrule
    &\multicolumn{4}{c}{\sc Ellipsis}\\
    \midrule
    Sluice Ellipsis & 1.4k & 480 & 992 & 351 \\
    VP Ellipsis & 264 & 20 & 78 & 984 \\
    \midrule
        &\multicolumn{4}{c}{\sc Auxiliary}\\
    \midrule
    OntoNotes & 153k & 18.8k & 19.5k & 463 \\
    WikiCoref & 5.6k & 630 & 638 & 2.2k \\
    SQuAD & 87.6k& 10.6k&-&117\\
    \bottomrule
    \end{tabular}
    \caption{QA pair counts and average context lengths (ACL) for different datasets, after conversion}
    \label{tbl:dataset-sizes}
\end{table}

\paragraph{Data Conversion}\label{qa-conversion} For converting the various datasets into the QA format of \texttt{<context, question, answer>} triples, we perform a simple restructuring as shown in Figure \ref{fig:qa_example}. We consider the entire document as the context; the sentence in which the ellipsis/mention is present becomes the question, and the antecedent/entity becomes the answer. In case of coreference resolution, where a single sentence can have $n$ mentions, we create $n$ questions where every question is the same sentence with a different mention $i\in \{1 \dots  n\}$ marked for resolution with \texttt{<ref>} and \texttt{</ref>} tags. Table \ref{tbl:dataset-sizes} shows the number of QA pairs created from each dataset and the average number of words in their contexts.

\paragraph{QA Architectures}
Generally, QA models have two main components: (i) an encoder module which learns to represent the question and its context, and (ii) a span selection module which predicts the start and end span indices of the answer if it is present in the context. In this work, we present experiments with three diverse models which take entirely different approaches to build the encoder module: (i) DrQA \cite{drqa}, with an LSTM  encoder, (ii) QANet \cite{qanet}, with a CNN encoder, and (iii) BERT \cite{bert}, with a (pretrained) transformer encoder. We use the three different models to show that the between-task synergies are relatively robust across architectures; but one architecture (BERT) is clearly superior to the others and will be the standard baseline we propose for future research.\footnote{Note that there are many differences between these architectures; not only the encoder networks. The number of parameters differ, and BERT is pre-trained on large volumes of data. Our purpose here is not comparing strategies, but simply showing that synergies can be seen across all architectures. For more details, see Appendix \ref{app:qa-models}.}



\begin{table*}
\centering
\small
\begin{tabular}{@{}r|l|lll|
lll@{}}
\toprule
\multicolumn{1}{c}{\sc{Task}} & \sc{SotA} & \multicolumn{3}{c}{\sc{Single Task}} & 
\multicolumn{3}{c}{\sc{Joint}} \\ \midrule
\multicolumn{1}{c}{\sc{}} & \sc{} & \sc{DrQA} & \sc{QANet} & \sc{BERT} 
& \sc{DrQA} & \sc{QANet} & \sc{BERT} \\
\midrule
\textbf{Sluice Ellipsis} & 70.00 \cite{sluice-short} & \textbf{77.48} & \textbf{75.70} & 
\underline{\textbf{85.10}} & \textbf{80.17} & \textbf{77.11} & \underline{\textbf{86.01}} \\
\textbf{VP Ellipsis} & 72.89 \cite{zhang-2019} & 62.86 & 1.93 & \textbf{76.42} & 
63.54 & 22.49 & \underline{\textbf{78.66}} \\
\end{tabular}
\caption{Ellipsis resolution scores are token-level F$_1$. Bold-faced results are better than the previous state-of-the-art; underlined results are the new state-of-the-art. When evaluated, our best joint architecture scores 72.31 on OntoNotes and 65.30 on WikiCoref (macro-averages of MUC, \bcub, and \ceaf scores). See Appendix \ref{sec:coref-compat} for why these numbers are not directly comparable to previously reported coreference resolution results in literature.}
\label{tbl:main-results}
\end{table*}

\begin{figure*}
    \centering
    \includegraphics[width=\textwidth]{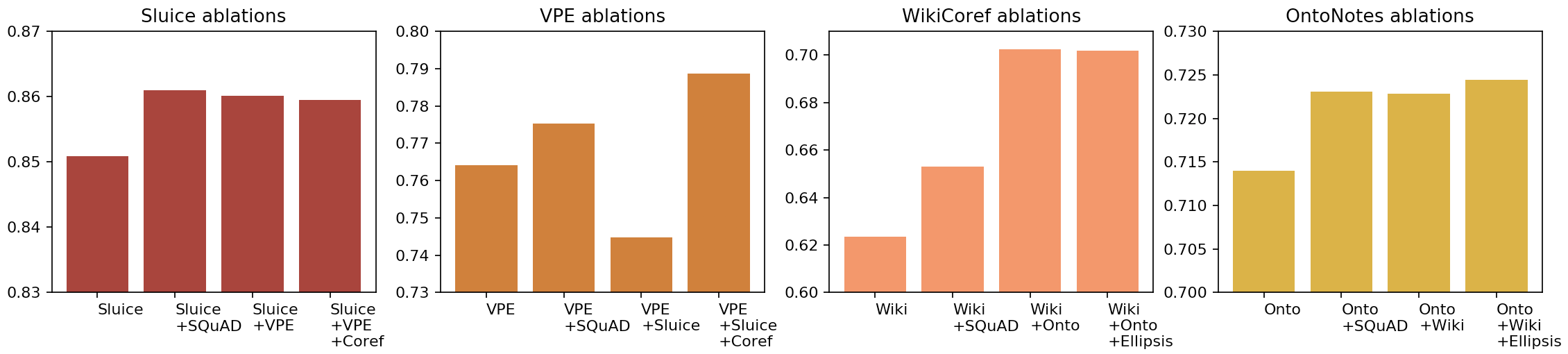}
    \caption{Dataset ablations (F$_1$)}
    \label{fig:ablation_plots}
\end{figure*}

\section{Experiments \& Results}
We conduct two sets of experiments: (i) the \textsc{Single-Task} experiments, in which we train and evaluate separate models for the two ellipsis resolution tasks; and (ii) the {\sc Joint} modelling experiments, where we train on the best possible combination of ellipsis resolution, coreference resolution and QA data, as determined on the validation set. The results can be seen in Table \ref{tbl:main-results}.\footnote{The reported results are the average of three independent runs with different random seeds.}

\paragraph{Single-Task Setup}
The \textsc{single-task} DrQA model improves the state-of-the-art on sluice ellipsis by $7.48$ F$_1$. 
The \textsc{single-task} QANet model also improves the state-of-the-art on sluice ellipsis by $5.7$ F$_1$, but fails to learn anything meaningful for VP ellipsis. We hypothesise this is due to the fact that $264$ training examples are not enough to train the model's large stack of encoder blocks from scratch. 

The \textsc{single-task} BERT model achieves state-of-the-art results in both the ellipsis datasets with absolute error reductions of $50.33\%$ (Sluice Ellipsis) and $13.02\%$ (VP Ellipsis). Interestingly, it also achieves a $17.10\%$ error reduction over the best previously reported results on WikiCoref, but see Appendix \ref{sec:coref-compat} for why such a direct comparison of numbers is not entirely fair.

\paragraph{Joint Setup} 
The {\sc Joint} models always perform on-par with, or better than the \textsc{Single-Task} models. In this setup, the BERT models beat the previous state-of-the-art for both Sluice and VP Ellipsis with $53.37\%$ and $21.28\%$ absolute error reductions respectively.

\begin{figure*}
    \centering
    \includegraphics[width=\textwidth]{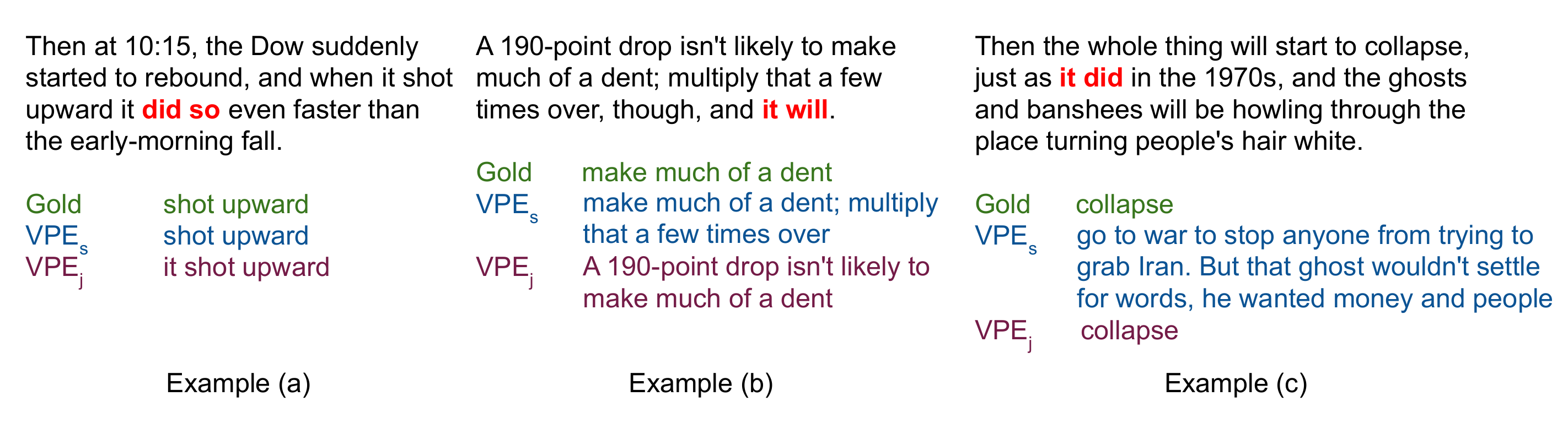}
    \caption{Selected gold and predicted antecedent spans from {\sc Single-Task} Verb Phrase Ellipsis (VPE$_s$ in figure) and {\sc joint} Verb Phrase Ellipsis (VPE$_j$ in figure) models.}
    \label{fig:VPE_analysis}
\end{figure*}

\section{Dataset ablations}\label{sec:ablations}

We determine the best task combinations on held-out validation data for each ellipsis resolution task.\footnote{These ablations are performed on the best performing (BERT) model.} For Sluice Ellipsis, the best results are obtained by training the models on a combination of Sluice and VP Ellipsis data. For VP Ellipsis, the best performance is attained when the models are trained with a combination of all datasets. When training a model for a particular task, we sample auxiliary data from other datasets to match the size of the main task's dataset. For each dataset, the variations in its F$_1$ scores of the best performing architecture when combined with other datasets are shown in Figure \ref{fig:ablation_plots}. The most interesting findings from these ablations are mentioned below. 

When the two ellipsis datasets are combined, the overall performance of the models increase for both tasks by around $1\%$ each. This shows that the two types of ellipsis are similar, and that when learning ellipsis resolution models, there is considerable synergy between the two resources. If we add subsampled coreference data when training these models, the Verb Phrase Ellipsis models gain up to $2.9$\%. One possible explanation could be more similarities between noun phrases and verb phrases, than between noun phrases and the sentences that are elided in Sluice Ellipsis resolution.

\section{Error Analysis}

We now look at some errors made by our best performing models. First, we compare the errors made by our \textsc{Single-Task} and \textsc{Joint} Sluice Ellipsis resolution models before moving on to VP Ellipsis.\footnote{We also briefly discuss how coreference resolution benefits from synergies with ellipsis in Appendix \ref{sec:coref-analysis}.}

\paragraph{Sluice Ellipsis}
The {\sc Joint} Sluice Ellipsis results improve modestly over the {\sc Single-Task} Sluice Ellipsis results. This is noteworthy, since the added VP Ellipsis data is quite small compared to the size of the sluice data. These models consistently select an antecedent of the right syntactic form, which is normally a complete sentence. Many of the errors consist of empty outputs: {\sc Single-Task} Sluice Ellipsis produces 58 empty outputs, while {\sc Joint} Sluice Ellipsis produces 63. Another source of error is discontiguous antecedents. It is not unusual for the gold antecedent to be a discontiguous span \cite{Donecker:96}, but our models are not permitted to produce such antecedents, so these cases will always be a source of error. 

All the systems have problems when the antecedent follows the ellipsis, as in the following example: {\em I don't know why, but they seem to need a story.} We also compared the right and left periphery scores of sluices, and found better results predicting the right periphery: for {\sc Single-Task} Sluice Ellipsis, there were 678 matches on the left edge, and 733 on the right edge; for {\sc Joint} Sluice Ellipsis, there were 703 left matches and 734 right matches. 

\paragraph{Verb Phrase Ellipsis}
The {\sc Single-Task} VP models trained with just VP Ellipsis data improves on the current state of the art, and further improvement is observed when trained on auxiliary data, especially the Sluice Ellipsis resolution dataset. While the {\sc Joint} VP Ellipsis model is generally better than the {\sc Single-Task} model, joint training with Sluice Ellipsis resolution data also seems to introduce unfortunate biases. While the {\sc Single-Task} model always selects antecedents of the right syntactic form, i.e., verb phrases, the {\sc Joint}~model may select sentential antecedents. 
See examples in Figure \ref{fig:VPE_analysis}. 

In Example (a), the {\sc Joint} VP model incorrectly includes the subject {\it it}, presumably because the sluice data includes complete sentences as antecedents. Similarly in Example (b) -- though the {\sc Single-Task} model correctly chooses an antecedent beginning with the verb {\it make}, it continues with additional material that does not form a coherent antecedent. The {\sc Joint} result is also incorrect, but note that it consists of the complete sentence containing the correct VP antecedent. Example (b) presents the advantages and disadvantages of the joint ellipsis training data. While the two types of ellipsis require antecedents of different forms, they have similar requirements in terms of where in the context the antecedent is to be found. Example (c) further supports this point. Here the {\sc Joint} result is perfect, while the {\sc Single-Task} result finds an antecedent that is in the wrong part of the discourse. The {\sc Single-Task} model is slightly better with left periphery matches than right: we found 58 left and 55 right matches. This is reversed with the {\sc Joint} model, with 54 left and 60 right matches. 

\section{Related Work}


We are not the first to use question answering to redefine a set of tasks. Recently,~\newcite{he2015question} showed that semantic role labeling annotations could be solicited by asking simple questions that implicitly target predicate-argument relations in a sentence. Parallel to our work, \newcite{hou2020bridging} cast bridging anaphora resolution as question answering based on context. \newcite{wu-etal-2020-corefqa} and \newcite{li-etal-2020-unified} also reformulate coreference resolution and named entity recognition as QA. In the realm of re-framing relation extraction as a QA problem, \newcite{levy-re} and \newcite{xwikire} create monolingual and multilingual template based QA datasets respectively, which yield relation extraction models which were better at generalizing in the zero-shot setting. Extending this idea, \newcite{mccann2018natural} introduced the DecaNLP challenge, which casts 10 core tasks in NLP as question-answering problems. Similar to our work, their architecture jointly learns across all of these tasks. DecaNLP includes pronoun resolution, a subset of coreference resolution, but it does so only on a small, hand-crafted dataset; it does not address ellipsis.

\paragraph{Limitations of our approach} One limitation of our approach is that, like most previous work, we assume ellipsis and coreference resolution amount to finding antecedent spans that corefer with the target mention. This is not always the case; the elided material can: (i) have extra-linguistic antecedents, and (ii) refer to something that is contextually implied.  




\section{Conclusion} We present strong models for Sluice and Verb Phrase ellipsis resolution problems, by reformulating them as machine reading comprehension problems, significantly outperforming the previously best reported results. We also empirically show that training these models jointly and with auxiliary data from coreference resolution and question-answering further improves their performance.
Our code is publicly available at \href{https://github.com/rahular/ellipsis-baselines}{https://github.com/rahular/ellipsis-baselines}.


\section{Acknowledgements}
We thank the reviewers for their valuable feedback. Rahul Aralikatte and Anders S{\o}gaard are funded by
a Google Focused Research Award.

\bibliographystyle{acl_natbib}
\bibliography{ref}

\appendix

\section{Similarity between Ellipsis and Coreference Resolution}
Linguists have long pointed out deep links among different forms of ellipsis, as well as between ellipsis and pronominal anaphora. For example, \newcite{merchant2001syntax} presents a unified account of ellipsis phenomena within a minimalist syntactic framework, and theorists such as \newcite{postal1966so} and \newcite{elbourne2013definite} go so far as to argue that pronouns are also elliptical forms. The exact nature of the connections between ellipsis and anaphoric constructions remains a subject of controversy among linguists. However it is clear that there are rooted connections, and in our view these connections represent potential areas to be exploited with forms of knowledge transfer among datasets of different types.

Typically in NLP, ellipsis and coreference have been treated as distinct tasks. Possible exceptions include \newcite{Lin:ea:16}, who present a rule-based, feature-rich system for handling ellipsis and coreference in Chinese medical dialogues, but the synergy between the two subsystems is limited; and \newcite{Banjade:ea:15}, who reduce ellipsis and coreference to problems of alignment to an auxiliary text implicitly describing the universe of the dialogue in question. 

\section{QA Models}\label{app:qa-models}
We briefly describe the architectures of the QA models below. All experiments are conducted on a single 12 GB GPU. For all models, we use the hyperparameter values recommended in their respective papers.

\paragraph{DrQA}
The Document Reader component of DrQA consists of a context and a question encoder followed by two span prediction classifiers. The context encoder is a multi-layer bi-directional LSTM \cite{lstm} which takes in word embeddings \cite[][GloVe]{glove}, similarity based features (whether the token appears in the question in it's original, lowercase or lemma form), and other token level features (positional tags, named entities and term frequency) as input. The concatenation of each layer's hidden units is used as the context vector. The question encoder is another LSTM which takes word embeddings as input and combines the resulting hidden units using a simple attention mechanism to form the question vector. A bilinear term which captures the similarities between context and question vectors is used to combine the two vectors and the resulting vector used as input to the span prediction classifiers. The two classifiers predict the start and the end span respectively and are trained independently.

\paragraph{QANet}
In QANet, each encoder layer is a stack of depthwise separable convolutions followed by a multi-head self-attention mechanism placed inside a residual block.  Initially, words in the context and question are embedded using a combination of GloVe and character embeddings. They are then contextualized individually with an encoder block. The representations are then passed through a context-query attention layer to obtain a combined representation of the context and question. This is further passed through three encoding blocks before feeding it into a classifier for predicting the answer spans.


\paragraph{BERT}
We use the pre-trained BERT$_{BASE}$ uncased model to encode questions and their contexts. It has $12$ Transformer blocks, $12$ self-attention heads, and a hidden size of $768$. Word piece tokenization \cite{googlenmt} is performed, both on the context paragraph and the question. The boundaries of the two sequences are marked by dummy symbols. The context and the question are joined with a $[SEP]$ token in between, and the $[CLS]$ token is prepended at the beginning to form the input. The representation of the $[CLS]$ token is fed into a single-layer MLP with $2$ outputs which is used to predict the span indices.\footnote{We use the implementation detailed in \newcite{huggingface}.}

\section{Coreference Resolution}
In this section, we analyse the best performing coreference models and discuss why they cannot be compared with other works in literature.

\subsection{Error Analysis}\label{sec:coref-analysis}
\begin{figure}
    \centering
    \includegraphics[width=0.97\columnwidth]{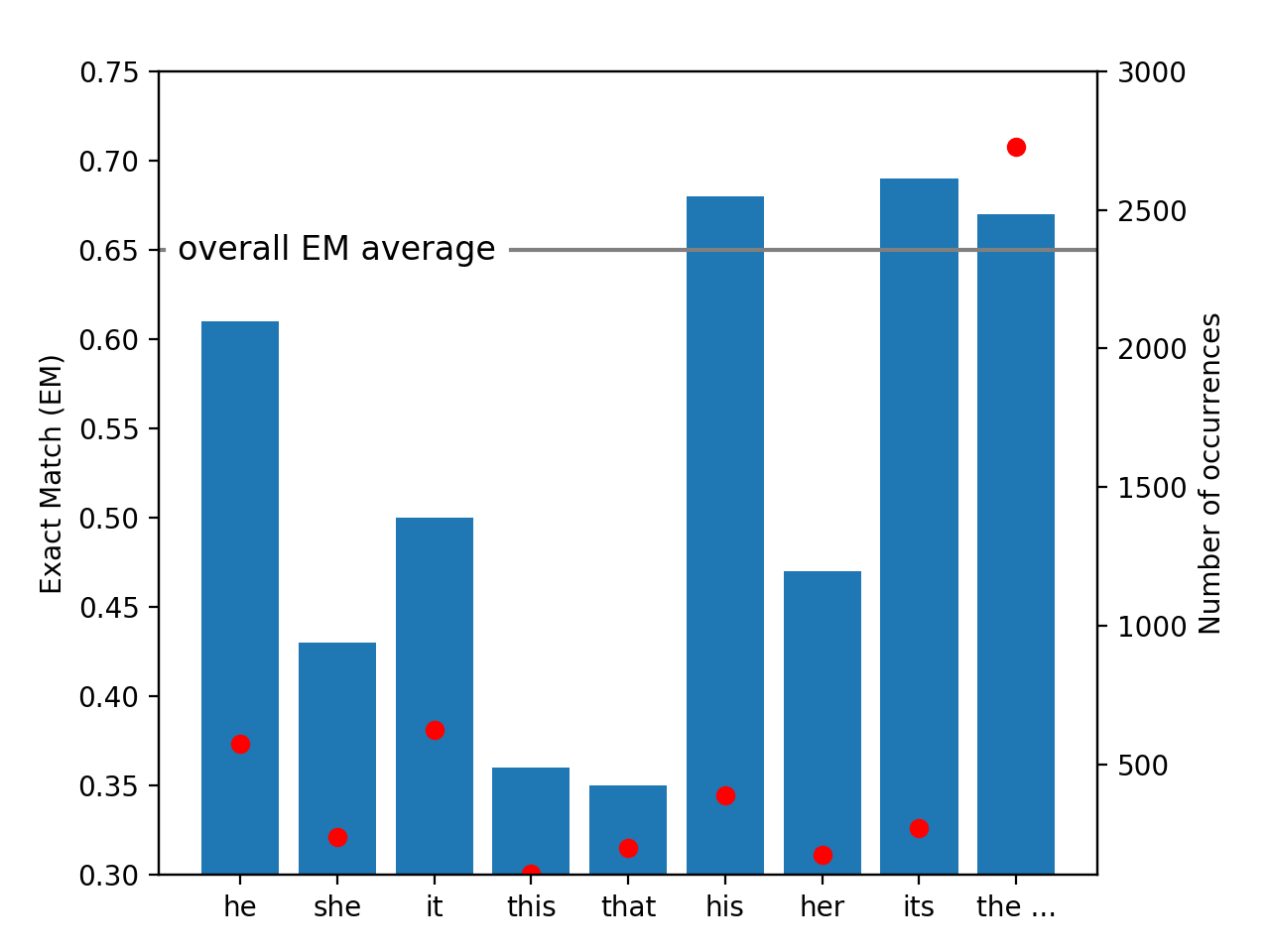}
    \caption{Exact match percentage (bars) and number of occurrences (dots) of referential forms in OntoNotes}
    \label{fig:ontonotes-analysis}
\end{figure}

The {\sc Joint} OntoNotes model improves a little over the {\sc Single-Task} counterpart. Here we examine specific referential forms in OntoNotes (WikiCoref has similar traits), as shown in Figure \ref{fig:ontonotes-analysis}. In general, performance is better on frequent pronouns -- e.g., `he' over `she', `this' and `that'. An exception to this is that `it' is less accurate, but more frequent than `he'. It is notable that the possessive pronouns (`his', `her', `its') are all more accurate than their nominative counterparts (`he', `she', `it'), perhaps because they tend to have a closer connection to their antecedents. Overall, the single-word referential forms are less accurate than multiple-word forms. For example, definite descriptions (forms beginning with `the') are more accurate than any of the single-word forms, with the exception of `its'. We speculate that multi-word forms provide more specific information, thus limiting the set of potential antecedents. It is also interesting to break down error by the grammatical gender of the pronouns. Male pronouns generally tend to be more accurate than their female counterparts. Antecedents of `he' and `his' are matched 20\% more frequently than for `she' and `her'. This is probably due to an unfortunate bias in OntoNotes, where female pronouns are 50\% rarer than male pronouns.

\subsection{Result Comparability}

\label{sec:coref-compat}
Converting coreference into QA fundamentally changes the coreference resolution problem: It, on the one hand, makes the coreference resolution problem harder, in that we require the identification of a specific antecedent span, rather than any mention in the entity chain; on the other hand, the problem becomes easier by providing the bracketing of the mention that needs to be resolved. Due to these differences, it is not possible to directly compare our results with others in literature. For analysis, to make our results more comparable with \newcite{lee2018}, we provided their model with the bracketing of the mentions and considered the first mention to be the antecedent. This way we can reinterpret their clusters as question-answer pairs and do not penalize them for mention bracketing errors, only considering pairs where they correctly identify mentions. Note this gives their model an advantage over ours, as their model considers multiple sources of evidence for inferring the coreference links, and gets to pick the subset of data on which the models are compared. On OntoNotes, in this setting, and after pruning around $7,358$ mentions \newcite{lee2018} bracketed wrongly, their new average F$_1$ score is $75.9$. Our performance on the same subset of the data is $72.1$. Upon manual inspection, we see the model in \newcite{lee2018} has a strong bias favoring nominal antecedents, whereas our model is more likely to predict clausal antecedents. On WikiCoref, our model remains better than the previous state of the art by some margin, with an F$_1$ of 69.2 over 43.6. 

\end{document}